\documentclass{article}
\usepackage{amsmath,graphicx,mlspconf}

\usepackage{booktabs} 
\usepackage{algorithm}
\usepackage{algorithmic}
\usepackage{amssymb}
\usepackage{hyperref}
\usepackage[capitalize,noabbrev]{cleveref}
\usepackage{enumitem}
\usepackage{color,soul}

\newcommand{\E}{\mathbb{E}}
\newcommand{\R}{\mathbb{R}}
\newcommand{\boldx}{\boldsymbol{x}}
\newcommand{\boldy}{\boldsymbol{y}}

\newcommand{\boldV}{\boldsymbol{V}}
\newcommand{\btheta}{\boldsymbol{\theta}}
\newcommand{\bLambda}{\boldsymbol{\Lambda}}
\newcommand{\calC}{\mathcal{C}}
\newcommand{\calD}{\mathcal{D}}
\newcommand{\calN}{\mathcal{N}}
\copyrightnotice{Preprint submitted to IEEE MLSP 2024}

\title{Leveraging Active Subspaces to Capture Epistemic Model Uncertainty in Deep Generative Models for Molecular Design}
\name{%
   A N M Nafiz Abeer$^{\star }$%
   \qquad Sanket Jantre$^{\dagger}$%
   \qquad Nathan M Urban$^{\dagger}$%
   \qquad Byung-Jun Yoon$^{\star \dagger}$
   \thanks{Funded by the ASCR program in the US DOE’s Office of Science under grant 0000269227  and  
projects B\&R\# KJ0402010 and FWP\# CC125}
}
\address{%
   $^{\star}$ Department of Electrical and Computer Engineering, Texas A\&M University, College Station, TX, USA \\%
   $^{\dagger}$ Computational Science Initiative, Brookhaven National Laboratory, Upton, NY, USA%
}

\begin{document}

\maketitle

\begin{abstract}
Deep generative models have been accelerating the inverse design process in material and drug design. Unlike their counterpart property predictors in typical molecular design frameworks, generative molecular design models have seen fewer efforts on uncertainty quantification (UQ) due to computational challenges in Bayesian inference posed by their large number of parameters. In this work, we focus on the junction-tree variational autoencoder (JT-VAE), a popular model for generative molecular design, and address this issue by leveraging the low dimensional active subspace (AS) to capture the uncertainty in the model parameters. Specifically, we approximate the posterior distribution over the active subspace parameters to estimate the epistemic model uncertainty in an extremely high dimensional parameter space.  
The proposed UQ scheme does not require any alteration of the model architecture, making it readily applicable to any pre-trained model. Our experiments demonstrate the efficacy of the AS-based UQ and its potential impact on molecular optimization by exploring the model diversity under epistemic uncertainty.
\end{abstract}
\vspace{-0.08in}
\section{Introduction}
\label{sec:intro}

Due to the advancement in deep learning, particularly in generative models, there has been increased application of generative design in material and drug discovery tasks. In small molecular drug design domain, different generative models \cite{gao2022sample, gomez2018automatic, jtvae_paper, jin20b_multi, olivecrona2017, MolDQN} are used for inverse design of molecules in conjunction with optimization strategies including Bayesian optimization, gradient ascent etc. In particular the generative molecular design (GMD) models like the Junction Tree Variational Autoencoder (JT-VAE) \cite{jtvae_paper} allow designing molecules with desired properties without directly optimizing over massive chemical space by embedding the molecules to a continuous and low dimensional latent space representation from high dimensional discrete chemical space.

Although the use of deep generative models in molecular design is on the rise, uncertainty of these large neural networks is relatively less explored in this domain. The existing works \cite{uq_prop_hirschfeld2020uncertainty, uq_prop_scalia2020evaluating, uq_prop_yang2023explainable, uq_prop_jiang2023uncertainty} primarily focus on the uncertainty of the classifier or regressor for molecular property prediction since the predicted properties of designed molecules often guide the inverse design process. Uncertainty quantification of the generative models that produce these molecules can lead to more robust application of the generative models in drug design. \cite{uq_opt_notin2021improving} demonstrated that epistemic uncertainty of the decoder in variational autoencoder (VAE) can enhance the efficiency of molecule generation task by enabling uncertainty aware optimization approach over the latent space.

Bayesian inference \cite{mackay1991bayesian, izmailov2021bayesian} technique is successfully used for capturing the uncertainty in predictions from the molecular property predictors \cite{uq_prop_scalia2020evaluating}. However, due to the large parameter space of generative molecular design models, e.g. JT-VAE (5.7M parameters), it is computationally challenging to learn the high dimensional posterior distribution over the generative model parameters.
Earlier efforts in Bayesian framework to address this issue trace back to the introduction of effective dimensionality of neural network parameter space in
\cite{mackay1991bayesian}. 
\cite{maddox2020rethinking} found that many directions in parameter space near local optima minimally impact neural network predictions. On one hand, \cite{frankle2018LOT} established that the heavily parameterized DNNs can be pruned leading to subnetworks with comparable performance. Accordingly, several pruning approaches in Bayesian deep learning lead to a substantially small subnetwork for tractable UQ \cite{blundell2015weight, Louizos-et-al-2017, jantre2023layer, jantre2023comprehensive}. On the other hand, inference over a low-dimensional subspace of the weights can effectively capture model uncertainty \cite{izmailov2020subspace, jantre2024active}. 
In this work, we proposed the use of active subspace \cite{jantre2024active} to quantify the epistemic uncertainty of JT-VAE model parameters where low dimensional active subspace facilitates the approximation of the posterior distribution using variational inference. 
Specifically, we use active subspace inference for its plug-n-play advantage over other subspace methods as we only need to collect small number of gradient samples from model weight perturbations to construct the subspace. 
Since the proposed approach does not require any change in the architecture of JT-VAE model unlike the dropout method for Bayesian approximation \cite{gal2016dropout}, it can be used as a plug-in tool for quantifying the uncertainty of GMD model in the established molecular design pipeline.

Our main contributions can be summarized as follows:
\begin{itemize}[noitemsep,nolistsep]
    \item We construct a low dimensional active subspace for capturing the epistemic uncertainty of a JT-VAE model which is otherwise computationally challenging due to its high-dimensional parameter space.
    \item Additionally, we  perform active subspace inference over all four components of JT-VAE parameters -- graph encoder, tree encoder, graph decoder, and tree decoder -- separately to demonstrate the effectiveness of our approach in improving the predictive UQ compared to the pre-trained JT-VAE model. 
    \item We further investigate the impact of our active subspace inference over JT-VAE on its generated molecules in terms of 6 different molecular properties of interest.
\end{itemize}


\section{Methodology}
\label{sec:methodology}
\subsection{Junction Tree Variational Autoencoder (JT-VAE)}
The Junction Tree Variational Autoencoder or JT-VAE \cite{jtvae_paper} is a variational autoencoder based generative model that is used for several molecular design applications \cite{gao2022sample}. This model utilizes two different representations of input molecule -- molecular graph and its corresponding junction tree. The graph encoder projects the molecular graph to a 28-dimensional latent representation. Similarly, we have 28-dimensional latent vectors for the junction tree encoded by the tree encoder. The molecular graph embedding and the junction tree embedding together form the 56-dimensional latent space of JT-VAE.

A two-stage procedure is followed to reconstruct the molecule from its latent space representation. First, the tree decoder decodes the junction tree embedding to a set of sequential decision rules needed to reconstruct the corresponding junction tree iteratively. Next, based on the reconstructed junction tree, the graph decoder utilizes the graph embedding component in the latent vector to build the molecular graph. The JT-VAE model is trained by minimizing the reconstruction loss for the junction tree and the molecular graph and the KL divergence loss between the posterior and prior distribution over the 56-dimensional latent space. The detailed description of the JT-VAE model architecture as well as the decoding process from latent vector can be found in \cite{jtvae_paper}.

\subsection{Quantifying Uncertainty of JT-VAE Parameters}
Our goal is to quantify the epistemic uncertainty of JT-VAE model, $p(\btheta \vert \mathcal{D})$ which indicates the uncertainty about the model parameters after it is trained with the training dataset $\mathcal{D}$. However, directly learning this distribution over the parameters is computationally challenging due to the high dimensionality of JT-VAE parameter space. In this work, we leverage the low dimensionality of active subspace to capture this uncertainty of JT-VAE. Specifically, we first construct the active subspace around the pre-trained JT-VAE model parameters followed by approximation of the posterior distribution over active subspace parameters. Finally, the samples from the learned posterior distribution are transformed to the original high dimensional parameter space of JT-VAE.

The quantified epistemic uncertainty can improve the predictive performance of JT-VAE. The pre-trained JT-VAE model parameters are the maximum likelihood estimate found by minimizing the training loss. Having a distribution over parameters instead of this point estimate helps the JT-VAE model to have better predictive uncertainty since it makes the model more robust to diverse set of training molecules.

\subsection{Active Subspace of Deep Neural Network}
\label{AS_summary}
\begin{figure}[!htb]
\begin{minipage}[b]{1.0\linewidth}
  \centering
  \centerline{\includegraphics[width = 0.45\textwidth]{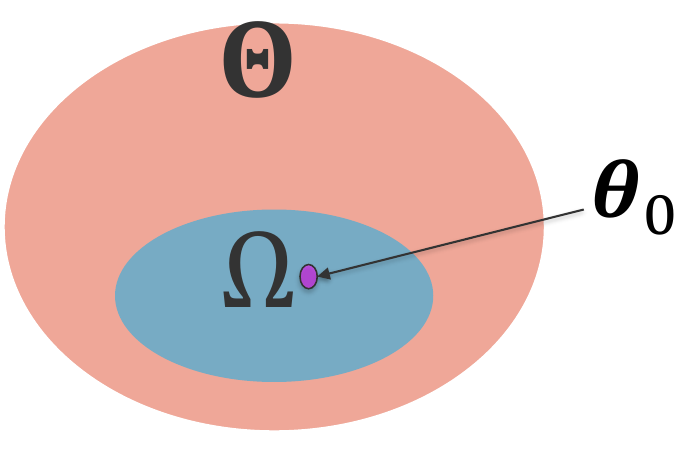}}
 \vspace{-0.2cm}
\end{minipage}
\caption{Low-dimensional active subspace (AS) $\boldsymbol{\Omega}$ around the pre-trained DNN parameter $\btheta_0 \in \boldsymbol{\Theta}$. $\btheta_0$ is the maximum likelihood estimate obtained by minimizing the training loss.}
\label{fig:active_sub}
\vskip -0.2in
\end{figure}

In order to reduce the high dimensionality of a DNN model, we employ the active subspace (AS) of the model parameters similar to \cite{jantre2024active}. We identify a low-dimensional subspace, $\boldsymbol{\Omega}$ within a high-dimensional parameter space $\boldsymbol{\Theta}$ (\cref{fig:active_sub}) that most significantly influences the variability in the DNN's output on average. First we consider a continuous function $f_{\btheta}(\boldx)$ with $\boldx$ and $\btheta \in \R^D$ denoting the input and the model parameters respectively. The parameters are stochastic with $\btheta \sim p(\btheta)$ being their probability distribution. We then construct a $D\times D$ uncentered covariance matrix of the gradient of $f_{\btheta}(\boldx)$: $\calC = \E_{\btheta}\left[(\nabla_{\btheta} f_{\btheta}(\boldx)) (\nabla_{\btheta} f_{\btheta}(\boldx))^T\right]$ which can be approximated using Monte Carlo sampling as follows:
\begin{equation}
    \hat{\calC} = \frac{1}{n} \sum_{i=1}^n (\nabla_{\btheta_i} f_{\btheta_i}(\boldx)) (\nabla_{\btheta_i} f_{\btheta_i}(\boldx))^T , \enskip \enskip \btheta_i \sim p(\btheta) \label{covar_apprx}
\end{equation}
where we generally have number of parameters, $D \gg n$. Now suppose $\hat{\calC}$ follows the eigendecomposition: $\hat{\calC} = \boldV \bLambda \boldV^T$ where $\boldV$ contains all the eigenvectors and $\bLambda=\rm{diag}(\lambda_1, \dots, \lambda_D)$ are the eigenvalues with $\lambda_1 \ge \dots \lambda_D \ge 0$. To extract subspace with most influential active directions, we split $\boldV$ in two parts $[\boldV_1, \boldV_2]$ where $\boldV_1 \in \R^{D\times k}$ and $\boldV_2 \in \R^{D \times (D-k)}$ with $k \le n \ll D$. The subspace spanned by $\boldV_1$ corresponds to the $k$ largest eigenvalues in $\bLambda$ and is considered as the ``active'' subspace. 
Accordingly, the projection matrix to reconstruct the model parameters from the subspace parameters is $\mathbf{P}=\boldV_1$.

\subsection{Active Subspace Construction for JT-VAE}
\label{sec:as_for_jtvae}
Let's partition high dimensional parameter space $\boldsymbol{\Theta}$ of JT-VAE into two non-overlapping regions: $\boldsymbol{\Theta}^D$ which contains the deterministic weights $\btheta^D$, and $\boldsymbol{\Theta}^S$ where we have the stochastic parameters, $\btheta^S$. When the entire parameter space $\boldsymbol{\Theta}$ is considered to be the deterministic partition, then we have the JT-VAE with deterministic parameters. On the other end, one can also consider all the parameters to be stochastic.

Our goal is to approximate the posterior distribution for the stochastic parameters, $\btheta^S\in \boldsymbol{\Theta}^S$ which can be still high dimensional. For example, if we select the tree decoder's parameters (2756131 in total) to be in the stochastic partition, then we have to approximate the 2756131-dimensional posterior distribution. Instead of directly approximating the high dimensional posterior distribution for $\btheta^S$, we construct a low dimensional active subspace $\boldsymbol{\Omega}$ within $\boldsymbol{\Theta}^S$ while keeping the deterministic parameters frozen at their pre-trained weights, i.e. $\btheta^D = \btheta_0^D$. In brief, we first randomly select 100 molecules from the training dataset of JT-VAE. For each of these molecules, we initialize the stochastic parameters $\btheta^S$ by drawing sample from $\mathcal{N}(\btheta_0^S,\sigma_0^2\mathbf{I})$, and the deterministic parameters $\btheta^D$ are set to their pre-trained weights, $\btheta_0^D$. With this initialized JT-VAE model, we compute the gradient of model loss corresponding to the molecule with respect to the stochastic parameters. This process is repeated for all 100 molecules, and the corresponding 100 gradient vectors help us to approximate uncentered covariance matrix, $\hat{\mathcal{C}}$ via \cref{covar_apprx}. We select the active subspace as the space spanned by the eigenvectors in $\mathbf{V}_1$ corresponding to $k$ largest eigenvalues of $\hat{\mathcal{C}}$. The projection matrix $\mathbf{P}=\mathbf{V}_1$ transforms the active subspace parameters, $\boldsymbol{\omega}\in \boldsymbol{\Omega}$ to the stochastic parameters space $\boldsymbol{\Theta}^S$ according to \cref{eq:projection}.
\begin{align}
    \btheta^S = \btheta_0^S + \mathbf{P}\boldsymbol{\omega} \label{eq:projection}
\end{align}

\cref{alg:AS} shows the procedure for active subspace inference over the stochastic parameters $\btheta^S$. This is the likelihood-informed AS \cite{jantre2024active} with $f_{\btheta}$ being the loss function $\mathcal{L}$ that is used for training the JT-VAE model which includes the reconstruction loss and KL-divergence loss of JT-VAE. 

\begin{algorithm}[!htb]
\caption{Active subspace inference for JT-VAE}
\label{alg:AS}
\begin{algorithmic}[1]
    \STATE {\bfseries Input:} loss function $\mathcal{L}$ used to train $\mathcal{M}_{\boldsymbol{\theta}_0}$, pre-trained model weights $\btheta_0 = \left[ \btheta^S_0, \btheta^D_0 \right]$, training dataset $\mathcal{D}$ of pre-trained model, number of gradient samples $n$, active subspace dimension $k$, perturbation standard deviation $\sigma_0$.
    \FOR {$j= 1,2,\dots, n$}
    \STATE Sample an input molecule $\boldx_j \in \mathcal{D}$
    \STATE Sample $\btheta^S_j \sim \calN(\btheta^S_0,\sigma_0^2\mathbf{I})$ 
    \STATE Compute $\nabla_{\btheta^S_j} \mathcal{L}(\mathcal{M}_{\boldsymbol{\theta}_j},\boldx_j)$ \hspace{2mm} where $\btheta_j = \left[ \btheta^S_j, \btheta^D_0 \right]$
    \ENDFOR
    \STATE Uncentered covariance matrix of loss gradients: \\ $\quad \hat{\calC} = \frac{1}{n} \sum_{j=1}^n (\nabla_{\btheta^S_j} \mathcal{L}(\mathcal{M}_{\boldsymbol{\theta}_j}, \boldx_j)) (\nabla_{\btheta^S_j} \mathcal{L}(\mathcal{M}_{\boldsymbol{\theta}_j},\boldx_j))^T$
    \STATE SVD decomposition: \\
    $\quad \hat{\calC} = \boldV \bLambda \boldV^T = [\boldV_1 \enskip \boldV_2] \begin{bmatrix} \bLambda_1 & 0 \\ 0 & \bLambda_2 \end{bmatrix} [\boldV_1 \enskip \boldV_2]^T $
    \STATE Projection matrix: $ \mathbf{P} = \boldV_1$ -- \textbf{active subspace} spanned by $\boldV_1$ corresponds with the $k$ largest eigenvalues in $\bLambda$
    \STATE Posterior distribution over corresponding subspace parameters $\boldsymbol{\omega}$ is learned using VI.
    \STATE Draw $M$ samples of active subspace parameters: $\boldsymbol{\omega}_m \sim p(\boldsymbol{\omega}|\calD) \quad \text{where,} \enskip m \in \{1,\cdots, M\}$  
    \STATE Compute JT-VAE model weights for each sample: $\btheta_m = \left[\btheta^S_0 + \mathbf{P} \boldsymbol{\omega}_m, \btheta^D_0 \right]$
\end{algorithmic}
\end{algorithm}

\subsection{Bayesian Framework}
The training data $\calD=\{(\boldx_j,\boldy_j)\}_{j=1,\cdots,N}$ consists of $N$ i.i.d. observations of inputs--$\boldx$ and outputs--$\boldy$. With prior distribution $p(\btheta)$, we infer the posterior distribution of $\btheta$ using the Bayes' rule: $p(\btheta|\calD) \propto p(\calD|\btheta) p(\btheta)$. With that, we can predict the output for new data point $\boldx^*$ through posterior marginalisation:
\begin{align*}
p(\boldy^*|\boldx^*,\calD) &= \int_{\btheta} p(\boldy^*|\boldx^*,\btheta) p(\btheta|\calD) d \btheta 
\end{align*}
Since the posterior $p(\btheta|\calD)$ is intractable, we use posterior approximation techniques-- specifically variational inference (VI) leveraging deterministic optimization to speed up the inference and scalability to large data \cite{Blei2017}. It infers the parameters of a distribution $q(\btheta)$ that minimises the Kullback-Leibler (KL) divergence of $q(\btheta)$ from the exact posterior $p(\btheta|\calD)$.
\vspace{-0.1in}

\subsubsection{Variational Inference for Posterior Approximation}
\label{sec:vi_training}
To approximate the posterior distribution $p(\boldsymbol{\omega}\vert \mathcal{D})$ over the $k$ dimensional active subspace parameters, we follow the mean field variational inference approach where the posterior is approximated by multivariate normal distribution with mean parameters, $\boldsymbol{\mu}_\text{post}$, and standard deviation parameters, $\boldsymbol{\sigma}_\text{post}$. We used the training dataset of JT-VAE to 
approximate the posterior distribution parameters -- $\boldsymbol{\mu}_\text{post}$ and $\boldsymbol{\sigma}_\text{post}$.
For each batch of training data, we first draw active subspace parameters from the posterior distribution approximation. Then these active subspace parameters are used to initialize the stochastic parameters, $\btheta^S$ via \cref{eq:projection}, while the deterministic parameters are fixed at their pre-trained weights. We applied the Adam optimizer \cite{kingma2014adam} with a learning rate of $0.001$ to learn $\boldsymbol{\mu}_\text{post}$ and $\boldsymbol{\sigma}_\text{post}$ by minimizing the combined loss of JT-VAE training loss for the initialized JT-VAE (that includes the reconstruction loss and KL divergence of JT-VAE) with KL divergence loss between approximated posterior distribution, $p(\boldsymbol{\omega};\boldsymbol{\mu}_\text{post},\boldsymbol{\sigma}_\text{post})$ and prior distribution $p(\boldsymbol{\omega}; \boldsymbol{\mu}_\text{prior}, \boldsymbol{\sigma}_\text{prior})$. For the prior distribution over AS parameters, a multivariate normal distribution with zero mean and $5$ standard deviation is used as a diffuse prior.

\subsubsection{Active Subspace Inference for JT-VAE}
\label{sec:subsapce_inference}
After training with the help of variational inference to approximate $p(\boldsymbol{\omega}\vert \mathcal{D})$, we use it to collect $M$ samples of active subspace parameters, $\left\{\boldsymbol{\omega}_m\right\}_{m=1}^M$, and construct corresponding $M$ different JT-VAE models with parameters $\btheta_m = [\btheta_m^S, \btheta_m^D]$ where  $\btheta_m^S = \btheta_0^S +\mathbf{P}\boldsymbol{\omega}_m$ and $\btheta_m^D = \btheta_0^D$. For each test datapoint, we utilize these $M$ models to quantify uncertainty in predictions in  Bayesian model averaging fashion.

\section{Numerical Results and Discussion}
\label{sec:experiments}

\subsection{Numerical Experiment Details}

We constructed 20 dimensional active subspace with perturbation noise standard deviation, $\sigma_0=0.1$ for 5 different choices of stochastic and deterministic partitions of JT-VAE parameters. First, we considered all parameters of JT-VAE to be in stochastic partition $\boldsymbol{\Theta}^S$, this is denoted as AS of JT-VAE in \cref{nll-table}. 
In this case, 20 dimensional active subspace is utilized to capture the uncertainty for all 5664143 parameters of JT-VAE.
For the other four cases, we treat each subnetwork-- tree encoder, graph encoder, tree decoder and graph decoder respectively -- to be in stochastic partition, and consider the rest of the parameters as deterministic parameters. For example, in AS of JT-VAE tree encoder's case, all 1998056 parameters of the tree encoder are considered to be stochastic, and we construct the active subspace for these stochastic parameters while rest of the JT-VAE parameters are fixed at their pre-trained weights. The pre-trained JT-VAE from \cite{tripp2020sample} is the case when all of the parameters are deterministic.

We followed the procedure described in \cref{sec:as_for_jtvae} to learn the projection matrix $\mathbf{P}$  around the pre-trained model weights of the parameters in the stochastic partition. Then we approximate the posterior distribution, $p\left(\boldsymbol{\omega} \vert \calD\right)$, over the $20$-dimensional subspace parameters using variational inference described in \cref{sec:vi_training}. Finally we collect $M=10$ samples independently from the learned variational distribution over subspace parameters -- $p\left(\boldsymbol{\omega} ; \boldsymbol{\mu}_{\text{post}}, \boldsymbol{\sigma}_{\text{post}}\right)$ to perform active subspace inference.
We learn the active subspace and corresponding posterior distribution using the training dataset from \cite{tripp2020sample}.
\subsection{Improvement in Predictive Uncertainty Estimation}

To empirically investigate whether active subspace inference can improve the predictive uncertainty of the pre-trained JT-VAE model, we perform inference over the validation dataset in \cite{tripp2020sample} using pre-trained model weights as well as the models sampled from the learned active subspace posterior distributions.
For the pre-trained JT-VAE, we compute the negative log-likelihood (nLL) over the validation dataset $50$ times.
In case of active subspace inference, we apply \cref{eq:projection} to construct the corresponding JT-VAE models from 10 independent active subspace posterior samples. 
For each of these 10 models, nLL computation is repeated 5 times (to account for the sampling in latent embedding) over the same validation dataset. The number of repetitions is chosen this way so that we have $50$ predictions per sample for all cases.

\begin{table}[t]
\vskip -0.1in
\caption{Negative log-likelihood (nLL) comparison between pre-trained JT-VAE and different active subspace of JT-VAE}
\label{nll-table}

\begin{center}
\begin{small}
\begin{tabular}{lcr}
\toprule
\textbf{Inference type} & \textbf{nLL}  \\
\midrule
Pre-trained JT-VAE  &  1.4812 $\pm$ 0.003
\\
AS of JT-VAE  & 1.4599 $\pm$ 0.003 \\
AS of JT-VAE tree encoder  & \textbf{1.4596} $\pm$ \textbf{0.003}\\
AS of JT-VAE graph encoder  & 1.4975 $\pm$ 0.003\\
AS of JT-VAE tree decoder  & 1.4621 $\pm$ 0.003
\\
AS of JT-VAE graph decoder  & 1.4807 $\pm$ 0.003\\
\bottomrule
\end{tabular}
\end{small}
\end{center}
\vskip -0.35in
\end{table}

\begin{table*}[!htb]
\caption{Top 10\% average property values for 1,000 random latent points decoded by pre-trained JT-VAE and active subspace (AS) inference over entire JT-VAE as well as its decoders. For AS inference, average and standard deviation over 5 repetitions is reported and the ones with the largest change in top 10\% average property from the pre-trained JT-VAE are highlighted.}
\label{prop-table}

\begin{center}
\begin{small}
\vskip -0.2in
\begin{tabular}{lcccccc}
\toprule
\textbf{Inference type} & logP ($\uparrow$) & SAS ($\downarrow$) & NP score ($\uparrow$) & DRD2 ($\uparrow$) &JNK3 ($\uparrow$) & GSK3$\beta$ ($\uparrow$)\\
\midrule
Pre-trained JT-VAE  &  4.2542 & 2.0126 & 0.0377& 0.0484& 0.0580 & 0.1310
\\
AS of JT-VAE  & 4.2739 (0.0093) & 2.0112 (0.0011) & 0.0443 (0.0061)  & 0.0503 (0.0001) & \textbf{ 0.0605 (0.0002)} & 0.1375 (0.0002)\\
AS of tree decoder  & \textbf{4.2871 (0.0078)} & \textbf{2.0078 ( 0.0022)}& \textbf{0.0488  (0.0049)} & \textbf{0.0421 (0.0029)} &0.0596 (0.0003) & \textbf{0.1403 (0.0002)}\\
AS of graph decoder  & 4.2539 (0.0006) & 2.0147 (0.0001)&  0.0361 (0.0002) & 0.0486 (0.0000) &0.0583 (0.0001) & 0.1386 (0.0000) \\  
\bottomrule
\end{tabular}
\end{small}
\end{center}
\vskip -0.3in
\end{table*}
\cref{nll-table} shows the average and standard deviation of the nLL metric for different inference methods. The lower nLL value indicates better predictive performance for the corresponding inference type. Our result shows that the active subspace inference over all JT-VAE  parameters improves the pre-trained JT-VAE's capability to reconstruct the molecules. 
In fact, we get further improvement by adopting active subspace over only the tree encoder parameters.
The junction tree of the molecule (that the tree encoder embeds) plays a significant role in reconstructing the molecules in JT-VAE since we get the molecular graph using the decoded junction tree. And by constructing the active subspace over the tree encoder parameters we are allowing the posterior models more flexibility to learn the latent space representation of the diverse junction tree structures of the dataset. This is also indicated by a similar improvement in nLL for the AS over the tree decoder.
Whereas in the case of AS over all JT-VAE parameters, the posterior models still manage to achieve closer performance. This also highlights the effectiveness of active subspace in learning the subspace that is most significant to JT-VAE loss within its 5.7M parameter space. Interestingly, we get poor performance for active subspace over the graph encoder. This can be interpreted as the pre-trained graph encoder possibly being at a relatively sharper loss landscape region and the stochasticity of the approximated AS posterior might be pushing models towards unfavorable region.

\subsection{Impact on Molecular Inverse Design}

One of the popular approaches for designing molecules with desired properties includes optimization over the learned latent space of JT-VAE like generative models. Specifically, the authors in \cite{jtvae_paper} first trained a Gaussian process-based surrogate model to predict the properties of molecules from their corresponding latent space representation encoded by JT-VAE model. Then Bayesian optimization (BO) was performed iteratively over the latent space using the pre-trained JT-VAE model parameters to search for molecules with optimized properties. Now thanks to active subspace inference, we have multiple JT-VAE models drawn from the active subspace posterior. In this section, we investigate whether these models sampled from active subspace posterior act differently compared to the pre-trained JT-VAE during the reconstruction of molecules from the latent space.

Specifically, we consider a set of 1000 latent points drawn from the latent space of JT-VAE according to $\mathcal{N}(\mathbf{0}, \mathbf{I})$ which are deocoded by pre-trained JT-VAE, and we evaluate their properties using property predictors described in \cref{prop_interest}. 
Similarly, these 1000 latent points are processed by models sampled from active subspace posterior. We draw 10 samples from the active subspace posterior, and the same 1000 latent points are distributed uniformly to each of the corresponding models for decoding. 
For fair comparison with pre-trained JT-VAE's performance, we allow 100 latent points per sampled model so that we have 1000 decoded molecules at the end. \cref{prop-table} contains the average property for the top 10\% unique molecules out of 1000 latent points decoded by pre-trained JT-VAE and 3 selected types of AS inference -- the active subspace over all JT-VAE parameters, the tree decoder and the graph decoder. The cases of AS over tree encoder and graph encoder are excluded since we are reconstructing the molecules from the given latent vectors using the decoders. 
We infer about the diversity introduced by the posterior models by analyzing the change in properties of decoded molecules compared to the case when the same latent points are decoded by the pre-trained model. 
AS over tree decoder shows higher diversity of posterior models, demonstrated by the maximal change in top 10\% average property compared to pre-trained JT-VAE across 5 out of 6 properties. This is intuitive since the diversity in the decision rules  (learned by the tree decoder) for constructing the junction tree directly impacts the reconstructed molecules (and their properties). We have similar but reduced impact on the generated molecules for AS inference over all JT-VAE parameters as well as the graph decoder. 
The change for AS of graph decoder is small since it only decides the arrangement between molecular units of predicted junction tree from tree decoder and this has a lower impact on molecular properties.

\subsubsection{Properties of Interest}
\label{prop_interest}
\vskip -0.1in
We have considered six chemical properties: water-octanol partition coefficient (logP), synthetic accessibility score (SAS), natural product-likeness score (NP score) \cite{HARVEY2008894}, inhibition probability against Dopamine receptor D2 (DRD2) \cite{drd2_structure}, c-Jun N-terminal kinase-3 (JNK3) and glycogen synthase kinase-3 beta (GSK3$\beta$) \cite{li2018multi} which are often used as optimization objectives during the development of machine learning algorithms for drug design. For SAS, the top 10\% properties are the ones with lower property values, as lower SAS is desirable in a designed molecule. The property predictors for logP, SAS, and NP score \cite{NP_score_paper} are from RDKit package \cite{landrum2013rdkit}. For DRD2 inhibition probability, we used the support vector machine (SVM) classifier proposed in \cite{olivecrona2017}. We utilized the oracles from Therapeutics Data Commons \cite{Huang2021tdc} for prediction of inhibition probability for JNK3 and GSK3$\beta$. 

\section{Conclusion}
\label{sec:conclusion}
We demonstrated how leveraging a low dimensional active subspace of a deep generative model (specifically, JT-VAE in this study) can facilitate UQ, which would have been otherwise computationally intractable due to its huge parameter space. Our approach enables efficient UQ for a pre-trained JT-VAE model without requiring any modifications in the model architecture, making the technique readily applicable to existing molecular design pipelines utilizing such models. The AS inference over the entire JT-VAE as well as its four components reveals that the learned AS can effectively identify the subspace within the high-dimensional parameter space that has the largest influence on the JT-VAE loss. The identified model uncertainty class reflects a diverse pool of JT-VAE models that affect the molecules (and their properties) from the latent space, in a manner different from the pre-trained model. We expect that exploring the model diversity within the uncertainty class identified via AS has the potential to enhance the generative models' performance for specific downstream tasks in the molecular design process, in a manner complementary to existing model optimization schemes \cite{abeer2022multi}. 


\bibliographystyle{IEEEbib}
\bibliography{refs}

\end{document}